\def\year{2021}\relax
\documentclass[letterpaper]{article} 
\usepackage{aaai21}  
\usepackage{times}  
\usepackage{helvet} 
\usepackage{courier}  
\usepackage[hyphens]{url}  
\usepackage{graphicx} 
\usepackage{natbib}  
\usepackage{caption} 
\usepackage{subfigure}
\usepackage{amsmath,amssymb,amsfonts}
\usepackage{amsthm,amsmath,amssymb}
\usepackage{mathrsfs}
\usepackage{bm}
\usepackage{algorithmic}
\usepackage{booktabs}
\usepackage[switch]{lineno}

\title{Explainable Tensorized Neural Ordinary Differential Equations for Arbitrary-step Time Series Prediction}
\author{

    Penglei Gao, \textsuperscript{\rm 1}
    Xi Yang, \textsuperscript{\rm 2}
    Rui Zhang, \textsuperscript{\rm 1}
    Kaizhu Huang, \textsuperscript{\rm 2}
    \\
}
\affiliations{

    \textsuperscript{\rm 1}School of science, Xi'an Jiaotong Liverpool University\\
    \textsuperscript{\rm 1}School of Advanced Technology, Xi'an Jiaotong Liverpool University\\

    111 Renai Road, Suzhou Industrial Park\\
    Suzhou, Jiangsu Province, P. R. China 215123\\
 
}

\begin{document}

\maketitle

\begin{abstract}
We propose a continuous neural network architecture, termed Explainable Tensorized Neural Ordinary Differential Equations (ETN-ODE), for multi-step time series prediction at arbitrary time points.
Unlike the existing approaches, which mainly handle univariate time series for multi-step prediction or multivariate time series for single-step prediction, ETN-ODE could model multivariate time series for arbitrary-step prediction. In addition, it enjoys a tandem attention, w.r.t. temporal attention and variable attention, being able to provide explainable insights into the data.
Specifically, ETN-ODE combines an explainable Tensorized Gated Recurrent Unit (Tensorized GRU or TGRU) with Ordinary Differential Equations (ODE). The derivative of the latent states is parameterized with a neural network. This continuous-time ODE network enables a multi-step prediction at arbitrary time points.
We quantitatively and qualitatively demonstrate the effectiveness and the interpretability of ETN-ODE on five different multi-step prediction tasks and one arbitrary-step prediction task.
Extensive experiments show that ETN-ODE can lead to accurate predictions at arbitrary time points while attaining  best performance against the baseline methods in standard multi-step time series prediction.
\end{abstract}

\section{Introduction} \label{intro}
Multi-step time series prediction is a crucial topic which has significant impacts on daily life, such as social science, finance, engineering, and meteorology. In such fields, an effective decision often depends on accurate predictions for multiple future values on relevant time series data. 
In multivariate time series, different exogenous variables might make different contributions to the target series. These contributions could be weakened by the processing mode in deep neural networks, such as Recurrent Neural Networks (RNNs), which blends the information of all exogenous variables into non-transparent hidden states.
To further improve the prediction accuracy and promote applications in more demanding scenarios, recent advanced challenges are providing interpretable insights into the multivariate time series and building continuous networks for predicting multiple future values at arbitrary time points.

\begin{figure}[t] 
	\centering
	\subfigure[Multi-step time series prediction by discrete neural networks.]{
		\label{diff.sub1}
		\includegraphics[width=0.41\textwidth]{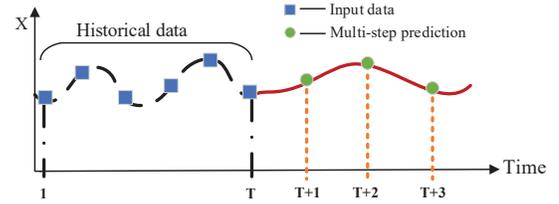}}
	\quad
	\subfigure[Arbitrary-step time series prediction by continuous neural networks.]{
		\label{diff.sub2}
		\includegraphics[width=0.41\textwidth]{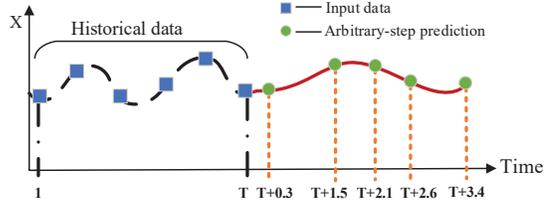}}
	\caption{(a) Discrete neural networks make predictions at time points $T+1$, $T+2$, and $T+3$. 
	(b) Continuous neural networks make predictions at arbitrary time points, e.g. $T+0.3$ or $T+2.6$.
	}
	\label{diff}
\end{figure}

Currently, the basic idea behind ‘interpretability’ is to develop an attention mechanism to distinguish the different contributions among different exogenous time series~\cite{guo2019exploring,heo2018uncertainty}. Although existing explainable methods achieved good prediction performance, they have certain limitations in multi-step prediction, e.g. discretizing observation and emission intervals would be required in building discrete neural networks. Fig.~\ref{diff} shows the different predictive ability between discrete neural networks and continuous neural networks.
For instance, the electricity consumption in industrial manufacturing is sampled hourly. Discrete neural networks could only make predictions at integer hour steps for multiple future values. The predicted time interval has the same sampled gap as the input data. In comparison, continuous neural networks overcome the limitation of integer-step prediction. They could forecast the next fifteen-minute value or the next forty-minute value when it is necessary even if the training data are hourly sampled. 

\begin{figure}[h]
\centering
\includegraphics[width=0.4\textwidth]{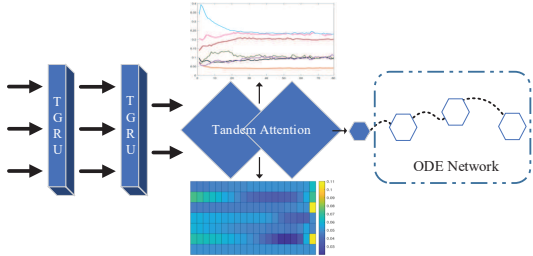}
\caption{Overview of the proposed ETN-ODE structure for arbitrary-step time series prediction.} \label{figOVER}
\end{figure}

In this work, we introduce a novel explainable continuous neural network for arbitrary-step time series prediction as shown in Fig.~\ref{figOVER}.
We involve the ordinary differential equations (ODE) in our framework to achieve continuous-time prediction by parameterizing the derivative of the latent states. The raw time series are encoded by a special designed Tensorized GRU. In addition, a tandem attention mechanism is designed to enhance the encoding ability and provide more adaptive input representation to the ODE network, which captures the different contributions among multivariate time series.

In multi-step time series prediction, modeling the incremental relationships could help to forecast the future values at arbitrary time steps with continuous neural networks. ODE is a kind of function describing the relationship of an unknown function and its derivative. This idea is beneficial for time series analysis since time is intrinsically continuous. Modeling the derivative in ODE makes the neural networks continuous when taking limits of the increment over time. If we could solve a function over time rather than a value, the arbitrary predicted values could be obtained with continuous forecasting steps. In many real-world applications, building continuous networks is sufficient to reduce the cost when high-frequency predictions are needed with low-frequency sampled input data.

To generate a better mapping of hidden states for the ODE network, we design a Tensorized GRU and a tandem attention to process the input multivariate times series.
First, the proposed TGRU modifies the internal structure of basic GRU cell to capture different dynamics of individual input features. We define the hidden state as a matrix rather than a vector at each time step. 
Then, the tandem attention w.r.t. temporal attention and variable attention is designed to provide interpretability into the input data and control information flow into the ODE network. 
Some exogenous series might have a significant influence on the target series with their short historical data while others with their long historical data. The temporal attention is utilized to capture the contributions in time aspect. As for variable attention, different contributions of exogenous series on the target series are captured representing different predictive power. 
At last, the tandem attention and hidden states are formed to produce a context vector as the input of a continuous-time ODE network to make predictions at arbitrary time points.
Ablation studies show the effectiveness of each component of our ETN-ODE architecture in the experimental section.

In summary, the major contributions are as follows:
\begin{itemize}
\item 
We propose a novel ETN-ODE framework, \textit{the first explainable continuous neural network typically designed for multivariate multi-step time series prediction at arbitrary time points} to our best knowledge,  which proves substantially more accurate than the current ODE model for arbitrary-step prediction in our experiments. 



\item 
We design a tandem attention to generate a more adaptive input to the  ODE network, offering further interpretability on the temporal and variable contribution to the target series when forecasting  future values at both the integer and continuous steps. 

\item We develop a Tensorized GRU  to process the multivariate time series for representing different dynamics of individual series.   Compared with the commonly-used Tensorized LSTM, the engaged Tensorized GRU enjoys much fewer parameters to be learned in the networks, which further improves the prediction performance. 


\end{itemize}

\section{Related Work} \label{RW}
\textbf{Time Series Predicting} Traditional methods such as Autoregressive (AR) model \cite{akaike1969fitting} and Vector Autoregressive model \cite{sims1980macroeconomics} cannot model non-linear relationships in multivariate time series, although they have shown their effectiveness for various real-world applications. 
In many real-world problems, it is helpful to predict multiple values given the historical information \cite{fox2018deep, yu2017long}. In the work of \cite{zang2017deep}, Zang et al. built a model with Long-Short Term Memory (LSTM) architecture to forecast multiple values on web traffic data.
\\
\textbf{Attention Mechanism} In addition to predicting, providing interpretability is also important for processing multivariate time series. Two types of attention generated with two independent RNNs are applied in~\cite{choi2016retain, heo2018uncertainty} to provide interpretable insights into the data. A mixture attention was proposed in~\cite{guo2019exploring} to enhance the interpretability with a tensorized LSTM structure for single-step prediction. The idea of tensorizing hidden states has shown its advantages for multivariate time series prediction in recent work~\cite{he2017wider, xutensorized}. The attention mechanism could help distinguish the different contributions among different exogenous time series.
\\
\textbf{Neural Networks with ODE} 
Recently, the Neural Ordinary Differential Equation~\cite{chen2018neural} proposed a continuous representation of neural networks which could produce predicting values at arbitrary time points for univariate time series. The ODE net is applied for irregularly-sampled time series classification in~\cite{rubanova2019latent}. They design an encoder with ODE-RNN to process the irregularly-sampled data instead of fixing the sampled gap by imputation. Another work of processing sporadically-observed time series in~\cite{de2019gru} combines Bayes method and ODE net for prediction. They utilize ODE net to impute the missing values to test the ability of reconstruction. In general, none of the existing continuous models deals with the multivariate time series for multi-step prediction which is conducted by our ETN-ODE model. We also combine the attention mechanism 
to provide interpretability for multivariate time series prediction.

\section{Main Method} \label{md}
\begin{figure}[]
\centering
\includegraphics[width=0.47\textwidth]{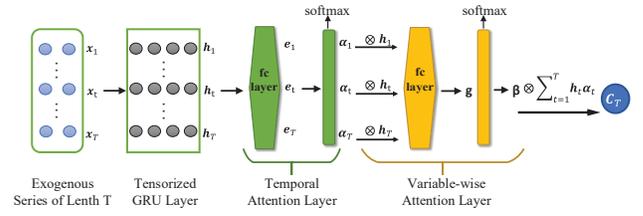}
\caption{Schematic of Tensorized GRU layer and tandem attention with two-variable input features and the hidden matrix of $2$-dimension per variable. The light blue circles represent the inputs, and the gray circles represent the hidden states. The green bar produces temporal attention $\bm{\alpha}$ given hidden states $\bm{h}$. The yellow bar produces variable attention $\bm{\beta}$ utilizing the information of hidden states and temporal attention. The dark blue circle represents the context vector $\bm{C}_T$ generated through $\bm{h}$, $\bm{\alpha}$, and $\bm{\beta}$.} \label{stru1}
\end{figure}

In this section, the details of our ETN-ODE model will be introduced. The model structure consists of Tensorized GRU, tandem attention, and ODE network. As shown in Fig.~\ref{stru1}, a Tensorized GRU is used to process the input multivariate time series. Two types of attention mechanisms are used to process the hidden states and generate a context vector $\bm{C}_T$. We leverage the context vector $\bm{C}_T$ to produce an initial value $\bm{z}_{T}$ that is assumed to have the Gaussian distribution with $[\bm{0}, \bm{I}]$.\footnote{Normal distribution with zero mean and identity variance.} Given the initial value and a predicted time interval, the ODE network could generate the multiple future values shown in Fig.~\ref{stru2}.

\subsection{Problem Statement} \label{PS}
Most of the deep learning models for time series prediction aim to forecast the future value of time $T+1$ given the historical data of previous $T$ time steps. In this paper, our target is to forecast arbitrary multiple future values with continuous time interval. 
Assume that we have a target series $\bm{Y}$ of length $T$ where $\bm{Y} = [y_1, y_2,\cdots,y_T] \in \mathbb{R}^T$. Given the predicted time interval $\{T+m_1,\cdots,T+m_K\}$, our aim is to design a proper non-linear mapping to forecast multiple future values of the target series representing as $\left[\hat{y}_{T+m_1},\cdots,\hat{y}_{T+m_K}\right] = \mathcal{F}(\bm{X}_T)$, where $K$ is the number of predicted values. The predicted time interval could be set as continuous values in testing stage for arbitrary-step prediction e.g. $\{T+0.3,\cdots,T+2.6\}$. Here $\mathcal{F}(\cdot)$ is achieved by an explainable continuous framework
and $\bm{X}_T$ is the multi-variable input data denoted as $\bm{X}_T = \{ \bm{x}_1,\bm{x}_2,\cdots,\bm{x}_T \}$, where $\bm{x}_t = [x^1_t,x^2_t,\cdots,x^N_t,y_t], t = 1,\cdots,T$. Thus, we will have $N+1$ input features for our model where $\bm{X}_T \in \mathbb{R}^{T\times({N+1})}$. The target series is related to $N$ exogenous time series, where $N$ is the number of exogenous series. Considering the auto-regressive effectiveness of the target series, we integrate $y_t$ and $N$ exogenous series at each time step of length $T$ as our multi-variable input features.

\subsection{Tensorized GRU} \label{GRU}
When processing with multivariate time series, we research on the internal structure of GRU cell to obtain a better mapping in hidden space. Here, we tensorize the hidden state 
to learn the independent representation of individual series based on the specific information from that time series. 
The hidden state is a matrix in our designed tensorized GRU rather than a vector in basic GRU at each step.
The matrix representation of hidden state involved more parameters. Tensorized GRU is more efficient, which only has an update gate and a reset gate compared to the Tensorized LSTM, leading to fewer parameters to be learned in networks. Results in ablation study also testify the effectiveness of Tensorized GRU.
The Tensorized GRU layer could be considered as a set of parallel GRUs, where each of them processes one individual series. 

To represent the tensorized GRU cell, we define the hidden state at time step $t$ as a matrix $\bm{H}_t = [\bm{h}^1_t,\cdots,\bm{h}^{N+1}_t]^\top$, where $\bm{H}_t \in \mathbb{R}^{(N+1)\times d}$, and $\bm{h}^n_t \in \mathbb{R}^d$ is the hidden state vector mapping the $n^{th}$ individual input feature at time step $t$, where $d$ is the hidden dimension of each input feature and $n = 1,\cdots,N+1$. The hidden size of GRU cell is derived as $D = (N+1) \times d$. According to the standard GRU networks, given the newly coming input data $\bm{x}_t$ and the previous hidden state matrix $\bm{H}_{t-1}$, the cell update is designed as:
\begin{equation}
\small
\begin{array}{cc}
    \bm{R}_t = \sigma(\bm{\mathcal{W}}_r \circledast \bm{H}_{t-1} + \bm{\mathcal{V}}_r \circledast \bm{x}_t + \bm{b}_r),&\\
    \bm{U}_t = \sigma(\bm{\mathcal{W}}_z \circledast \bm{H}_{t-1} + \bm{\mathcal{V}}_z \circledast \bm{x}_t + \bm{b}_z),&\\
    \tilde{\bm{H}}_t = tanh(\bm{\mathcal{W}}_h \circledast (\bm{R}_t \odot \bm{H}_{t-1}) + \bm{\mathcal{V}}_h \circledast \bm{x}_t + \bm{b}_h),&\\
    \bm{H}_t = (1 - \bm{U}_t) \odot \bm{H}_{t-1} + \bm{U}_t \odot \tilde{\bm{H}}_t, \label{eq.1}
\end{array}
\end{equation}
where $\bm{R}_t$ is the reset gate, $\bm{U}_t$ is the update gate, and $\tilde{\bm{H}}_T$ is the memory state. All of them have the same shape as the hidden state matrix. 
$\bm{\mathcal{W}}_*\in \mathbb{R}^{(N+1)\times d \times d}$ is the hidden-to-hidden transition tensor, and $\bm{\mathcal{V}}_*\in \mathbb{R}^{(N+1)\times d \times 1}$ is the input-to-hidden transition tensor. Terms $\bm{\mathcal{W}}_* \circledast \bm{H}_{t-1}$ and $\bm{\mathcal{V}}_* \circledast \bm{x}_t$ capture the information from the previous hidden state and newly coming input data respectively. Operation $\circledast$ means the product of two tensors along axis $(N+1)$, e.g. $\bm{\mathcal{W}}_r\circledast \bm{H}_{t-1} = [\bm{W}^1_
r\bm{h}^1_{t-1},\cdots,\bm{W}^{N+1}_r\bm{h}^{N+1}_{t-1}]^\top$, where $\bm{W}^n_r\bm{h}^n_{t-1} \in \mathbb{R}^d$. Operation $\odot$ denotes the element-wise multiplication. 
The produced hidden states would be sent to the tandem attention layer to generate the context vector as the input of ODE network.

\subsection{Tandem Attention}\label{Mix}
Two types of attention mechanisms are designed as the temporal attention and the variable attention to provide interpretability. Variable attention is generated based on the information distilled by temporal attention and hidden states considering the interaction between temporal attention and variable attention.
 Temporal attention is used to obtain the contributions in time aspect given the hidden states $\{\bm{h}^n_1,\cdots,\bm{h}^n_T\}$ corresponding to the specific $n^{th}$ input feature. The temporal attention weight $\alpha^n_t$ is computed as:
\begin{equation}
\small
    \alpha^n_t = \frac{exp(f_n(\bm{h}^n_t))}{\sum_{s}exp(f_n(\bm{h}^n_s))}, \label{eq.2}
\end{equation}
where $f_n(\cdot)$ could be a feedforward neural network specific to the $n^{th}$ input feature. The scalar $\alpha^n_t$ controls the influence of the $n^{th}$ input feature at time step $t$. Then a variable-wise attention layer is used to generate the contributions in feature aspect based on the temporal attention weight and hidden states. The variable attention weight $\bm{\beta}$ is computed as:
\begin{equation}
\small
    \bm{\beta} = softmax(f(\sum\nolimits_{t}\bm{\alpha}_t \odot \bm{H}_t)), \label{eq.3}
\end{equation}
where $f$ could be a feed-forward neural network and $\bm{\beta} \in \mathbb{R}^{N+1}$ controls the influence of each input feature for the prediction. Combining the two types of attention weights and hidden states, a context vector is generated as flows:
\begin{equation}
\small
    \bm{C}_T = \sum\nolimits_t\bm{\beta} \odot \sum\nolimits_{i=1}^d \bm{\alpha}_t \odot (\bm{H}_t)_i,  \label{eq.4}
\end{equation}
where $d$ is the hidden dimension per variable.

\subsection{Ordinary Differential Equation Network}\label{ode}
After the processing of tandem attention, the context vector $\bm{C}_T$ would be sent to the ODE solver providing outputs with continuous time intervals.
A hidden state in recurrent neural network decoder could be transformed as:
\begin{equation}
\small
    \bm{h}_{t+1}=\bm{h}_t + f(\bm{h}_t,\theta_t). \label{eq.5}
\end{equation}
When adding more layers and taking smaller steps in Eq.~\ref{eq.5}, parameterized continuous dynamics of hidden units could be obtained with an ordinary differential equation specified by a neural network in the limit \cite{chen2018neural}:
\begin{equation}
\small
    \frac{d\bm{h}(t)}{d(t)} = f(\bm{h}(t), t, \theta).  \label{eq.6}
\end{equation}
In this form, the ODE network uses neural networks to parameterize the derivative of the hidden states rather than directly parameterizing them, which leads to a continuous hierarchy instead of a discrete one. Furthermore, this mechanism requires a constant memory cost without storing any intermediate quantities of the forward pass during the training process. To get the numerical solution of $\bm{h}(t_1)$, the neural network $f(\bm{h}(t), t, \theta)$ could be integrated from $t_0$ to $t_1$ with the initial value $\bm{h}(t_0)$. The integral could be obtained through a black-box differential equation solver.
\begin{figure}[]
\centering
\includegraphics[width=0.47\textwidth]{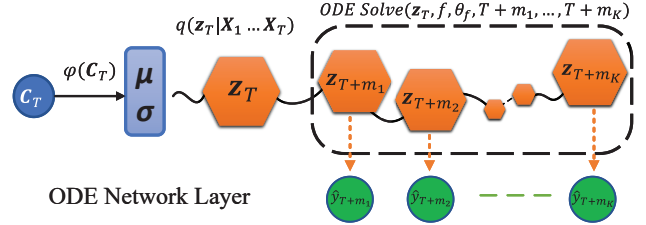}
\caption{An ODE network is applied to produce the predicted values given the initial value $\bm{z}_{T}$. $\bm{z}_{T}$ is sampled through a Gaussian distribution with $[\bm{\mu},\bm{\sigma}]$ that is generated by using  $\bm{C}_T$. The orange hexagon represents the latent states produced by the ODE solver. The green circles correspond to the predicted arbitrary-step values with a fully connected layer.} \label{stru2}
\end{figure}

In arbitrary-step prediction, given the predicted time interval $\{T+m_1,\cdots,T+m_K\}$, and an initial latent value $\bm{z}_{T}$, the ODE solver would produce the estimated latent states $\bm{z}_{T+m_1},\cdots,\bm{z}_{T+m_K}$. The ODE network is defined as:
\begin{equation}
\small
\begin{array}{cc}
    \bm{z}_{T} \sim p(\bm{z}_{T}), &\\
    \left[\bm{z}_{T+m_1},\cdots,\bm{z}_{T+m_K}\right] &\\
    = \mbox{ODESolver}(\bm{z}_{T}, f, \theta_f, {T+m_1},\cdots,{T+m_K}). \label{eq.7}
\end{array}
\end{equation}
In the black-box differential equation solver represented with a dotted rectangle in Fig.~\ref{stru2}, function $f$ is a time-invariant function that takes the value $\bm{z}_T$ at the current time step and outputs the gradient: $\partial \bm{z}(t)\diagup \partial t = f(\bm{z}(t), \theta_f)$, where $\theta_f$ are the parameters to be learned. Inspired by \cite{de2019gru}, we parameterize the function $f$ by using the GRU structure. With the property of the updated hidden state in GRU, a difference equation is obtained:
\begin{equation*}
\small
\begin{array}{cc}
         \Delta \bm{h}_t = \bm{h}_t - \bm{h}_{t-1} = (1 - \bm{u}_t) \odot \bm{h}_{t-1} + \bm{u}_t \odot \tilde{\bm{h}}_t - \bm{h}_{t-1} &\\
         \quad  = \bm{u}_t \odot (\tilde{\bm{h}}_t - \bm{h}_{t-1}).  \label{eq.8}
\end{array}
\end{equation*}
$ \Delta \bm{h}_t$ represents the increment between current hidden state and previous hidden state, while $\bm{u}_t$ and $\tilde{\bm{h}}_t$ are the update gate and memory state of the GRU cell. This difference equation naturally leads to the following ODE for $\bm{h}(t)$:
\begin{equation}
\small
    \frac{d\bm{h}(t)}{dt} = \bm{u}_t * (\tilde{\bm{h}}_t - \bm{h}_t). \label{eq.9}
\end{equation}
As observed from Fig.~\ref{stru2}, $p(\bm{z}_{T})$ is the prior distribution of $\bm{z}_{T}$ assumed as a Gaussian distribution which is approximated with the posterior distribution $q(\bm{z}_{T}|\bm{X}_1 \cdots \bm{X}_T)$. With the input features, we infer the parameters for the posterior distribution over $\bm{z}_{T}$:
\begin{equation}
\small
    q(\bm{z}_{T}|\bm{C}_T,\varphi) = \mathcal{N}(\bm{z}_{T}|\bm{\mu}_{\bm{z}_{T}},\bm{\sigma}_{\bm{z}_{T}}), \label{eq.10}
\end{equation}
where $\varphi$ are parameters to learn and $(\bm{\mu}_{\bm{z}_{T}},\bm{\sigma}_{\bm{z}_{T}})$ comes from the context vector $\bm{C}_T$ generated according to the TGRU layer and tandem attention layer. We would sample $\bm{z}_{T} \sim \mathcal{N}(\bm{\mu}_{\bm{z}_{T}},\bm{\sigma}_{\bm{z}_{T}})$. Given the arbitrary predicted time interval, model predictions of the target series could be obtained with the generated latent states $\bm{z}$ through the ODE network:
\begin{equation}
\small
    \left[\hat{y}_{T+m_1},\cdots,\hat{y}_{T+m_K}\right] = fc(\bm{z}_{T+m_1},\cdots,\bm{z}_{T+m_K}), \label{eq.11}
\end{equation}
where $fc$ represents the fully connected layer.

\subsection{Loss Function}\label{loss}
The objective function consists of three parts. The first part is the Mean Square Error (MSE) between the predicted values and the true values:
\begin{equation}
\small
    Loss_{mse} = \frac{1}{L}\sum\nolimits^L_{i=1} \frac{1}{K} \sum\nolimits^K_{t=1} (\hat{y}^i_t - y^i_t)^2, \label{eq.12}
\end{equation}
where $L$ is the number of training samples and $K$ is the number of predicted values. The second part is the KL-divergence between prior distribution $p(\bm{z}_{T})$ and posterior distribution $q(\bm{z}_{T})$:
\begin{equation}
\small
    Loss_{kl} = \frac{1}{L}\sum\nolimits^L_{i=1} D_{KL}(p(\bm{z}_{T})||q(\bm{z}_{T})), \label{eq.13}
\end{equation}
which is aiming to make them closer in latent space.
The third part is the negative log-likelihood of $\bm{y}$ describing the fitting degree of model parameters:
\begin{equation}
\small
    Loss_{nll} = -\frac{1}{L}\sum\nolimits^L_{i=1}\log p_\theta(\bm{y}^i), \label{eq.14}
\end{equation}
where $\log p_\theta(\cdot)$ is the logarithmic likelihood. Thus we have the loss function as:
\begin{equation}
\small
    Loss = Loss_{mse} + Loss_{kl} + Loss_{nll}. \label{eq.15}
\end{equation}
The goal of training is to minimize the loss function.

\section{Experiment and Evaluation}\label{exp}
We evaluate the proposed ETN-ODE model quantitatively and qualitatively in this section.

\subsection{Experimental Settings and Metrics} \label{setup}
We conduct our experiments on  four real-world datasets including SML2010 \cite{zamora2014line}\footnote{The room temperature is the target series and another $13$ attributes are selected as exogenous variables among $21$ attributes.}, PM2.5 \cite{liang2015assessing}, Energy \cite{candanedo2017data}, and NASDAQ100 \cite{qin2017dual}. More  details can be seen in the supplementary material.
For all the datasets, the train test split is set to $0.9$. The time window size $T$ for input data is set to $20$.

The proposed ETN-ODE model has been implemented using PyTorch $1.4$ and trained using the Adam algorithm with the mini-batch size $128$. The size of TGRU layers depends on the number of neurons per variable selected from $\{5, 10, 15, 20, 25\}$. $L2$ regularization is added with the coefficient selected from $\{0.001, 0.01, 0.1\}$. The noise standard deviation is chosen from $\{0.005, 0.01, 0.05, 0.1, 0.5, 1\}$. The learning rate is set to $0.01$. We train each method $5$ times and report the average performance for comparison. The evaluation has been performed on two standard predicted evaluation metrics, i.e. RMSE and MAE. 
\subsection{Baselines} \label{bl}
The five deep learning baselines are listed as below:\\
\textbf{Retain} \cite{choi2016retain}: This model uses two types of attention generated with two RNNs for classification with reversed time inputs. We replace the output layer with a fully connected layer for multi-step prediction.\\
\textbf{UA} \cite{heo2018uncertainty}: UA model uses variational inference on the attention part to learn the model uncertainty based on Retain. We replace the output layer with a decoder layer to adapt it for multi-step prediction.\\
\textbf{IMV-tensor} \cite{guo2019exploring}: The model designs a tensorized LSTM to capture different dynamics in multi-variable time series and mixture attention to model the generative process of the target series for the next one value prediction.\\
\textbf{TLSTM} \cite{yu2017long}: The model uses higher-order moments and higher-order state transition functions in LSTM for Long-term forecasting, using the encoder-decoder structure.\\
\textbf{Latent-ODE} \cite{chen2018neural}: The model uses a variational autoencoder structure with reversed time inputs. The encoder and decoder nets are both RNN networks.


\subsection{Evaluation Results}\label{result}

\subsubsection{Arbitrary-step Prediction}
\begin{table*}[h]
\small
\centering
\begin{tabular}{cc|cccc|cccc}
\hline\hline
           &         & \multicolumn{4}{c|}{RMSE}             & \multicolumn{4}{c}{MAE}              \\ \hline
Method     &         & SML2010 & Energy & PM2.5  & Nasdaq100 & SML2010 & Energy & PM2.5  & Nasdaq100 \\ \hline
           & Step1   & 0.099   & 76.552 & 29.661 & 11.559    & 0.079   & 36.829 & 18.127 & 9.838     \\
           & Step1.5 & 0.097   & 81.948 & 36.942 & 8.090     & 0.077   & 40.793 & 22.522 & 6.607     \\
ETN-ODE    & Step2   & 0.106   & 80.259 & 43.284 & 6.983     & 0.082   & 40.658 & 26.427 & 5.355     \\
           & Step2.5 & 0.118   & 84.237 & 48.159 & 6.950     & 0.089   & 42.425 & 29.803 & 5.397     \\
           & Step3   & 0.139   & 82.531 & 52.947 & 7.233     & 0.105   & 42.301 & 33.180  & 5.716     \\ \hline
           & Step1   & 0.647   & 84.134 & 51.600   & 33.997    & 0.557   & 43.871 & 38.759 & 30.515    \\
           & Step1.5 & 0.681   & 87.205 & 55.085 & 34.037    & 0.588   & 45.032 & 41.070  & 30.547    \\
Latent-ODE & Step2   & 0.718   & 85.532 & 59.116 & 34.085    & 0.621   & 45.041 & 43.755 & 30.595    \\
           & Step2.5 & 0.757   & 88.748 & 61.947 & 34.129    & 0.653   & 45.988 & 45.615 & 30.628    \\
           & Step3   & 0.799   & 86.993 & 65.391 & 34.178    & 0.685   & 45.907 & 47.731 & 30.676    \\ \hline\hline
\end{tabular}
\caption{Results of arbitrary-step prediction. A lower value corresponds to better performance.}
\label{arbitrary}
\end{table*}
Few studies focus on multivariate time series for arbitrary-step prediction by building continuous networks. Thus we only evaluate the effectiveness of arbitrary-step prediction of our proposed ETN-ODE against the Latent-ODE.
The intent is to forecast multiple future values which are not sampled in the original time series by setting continuous time intervals in testing stage based on Eq.~\ref{eq.7}.
To better demonstrate the arbitrary-step prediction quantitatively, we re-sample the dataset to half of its original size by taking twice the sampling gap.
During training, the model only outputs three future values at integer time points sharing the same sampled gap as the input data, e.g. $T+1, T+2, T+3$. While in the testing stage, the model would output extra two future values at continuous steps, e.g. $T+1.5$ and $T+2.5$, which are not involved during training.
For example, after re-sampling the energy dataset, we could forecast the values of the next `thirty-minute' and `fifty-minute' steps, when the data is sampled every twenty minutes.

\begin{figure}[b]
\centering
	\subfigure[Fitted values of `Step1.5']{
		\label{arbi.sub1}
		\includegraphics[width=0.21\textwidth]{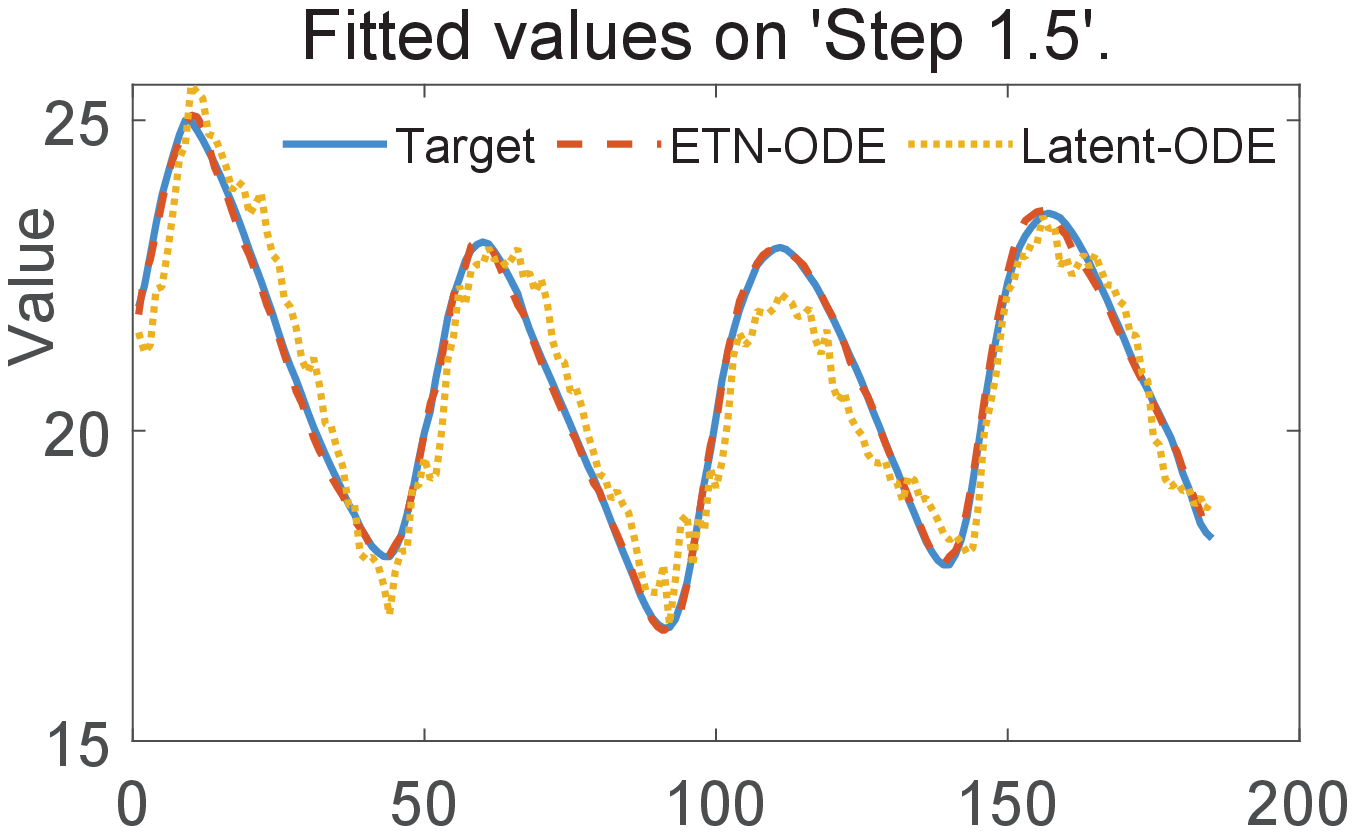}}
	\quad 
	\subfigure[Fitted values of `Step2.5']{
		\label{arbi.sub2}
		\includegraphics[width=0.21\textwidth]{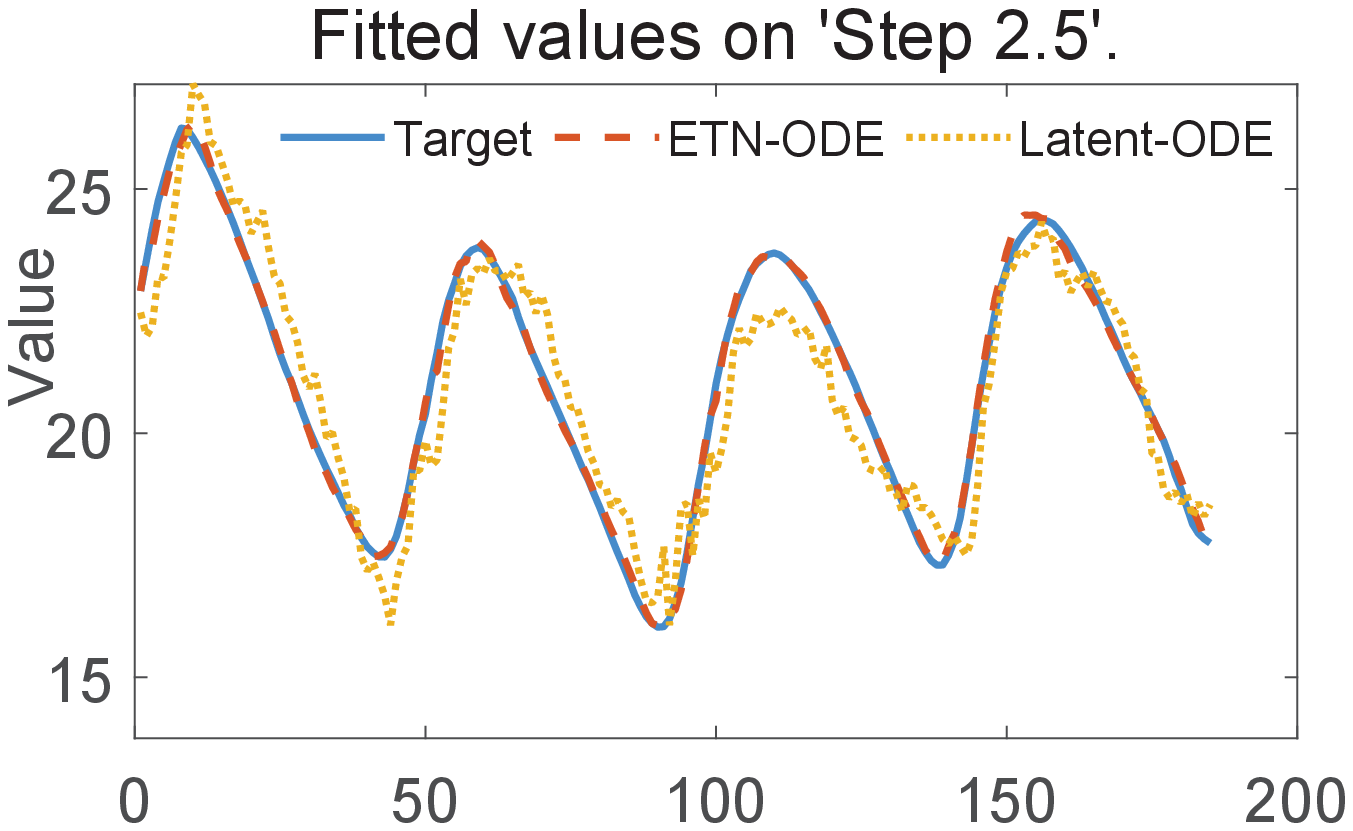}}
	\caption{Visualization of arbitrary-step prediction on SML2010.}
	\label{arbi}
\end{figure}

Table~\ref{arbitrary} shows the forecasting errors of each predicted step for the arbitrary-step prediction task. 
The results show that ETN-ODE can significantly outperform Latent-ODE on RMSE and MAE metrics due to the adaptability of the TGRU and tandem attention in multivariate time series forecasting.
`Step1.5' and `Step2.5' achieve similar forecasting errors to the other three steps demonstrating the effectiveness of our model for arbitrary-step prediction. In Fig.~\ref{arbi}, we further visualize the predicted values to the target series of `Step1.5' and `Step2.5' on SML2010.
The red dash line describes the predicted values of ETN-ODE, which captures the period information perfectly and fits much better in the target series than Latent-ODE. Additionally, it is noted that on Energy data, both the models lead to relatively larger errors, which can be partially explained by its highly complicated data nature. Nonetheless, ETN-ODE still generates better performance than  Latent-ODE. 
Overall, these results indicate the success of our ETN-ODE framework for multivariate time series prediction at arbitrary time points. 

\subsubsection{Standard Multi-step Prediction}
\begin{table*}[h]
\small
\centering
\begin{tabular}{c|ccc|ccc}
\hline\hline
Task       & K=1                   & K=5                   & K=10                  & K=1                   & K=5                   & K=10                  \\ \hline
Dataset    & \multicolumn{3}{c|}{SML2010}                                          & \multicolumn{3}{c}{Energy}                                           \\ \hline
TLSTM      & 0.926$\pm$0.006           & 0.974$\pm$0.099           & 1.022$\pm$0.021           & 81.312$\pm$1.038          & 90.783$\pm$2.617          & 93.258$\pm$1.999          \\
Latent-ODE & 1.193$\pm$0.020           & 0.538$\pm$0.036           & 0.631$\pm$0.048           & 78.589$\pm$1.030          & 84.801$\pm$0.598          & 86.135$\pm$0.849          \\
Retain     & 0.054$\pm$0.001           & 0.136$\pm$0.006           & 0.192$\pm$0.003           & 63.221$\pm$0.431          & 79.063$\pm$0.689          & 84.815$\pm$0.758          \\
UA         & 0.055$\pm$0.001           & 0.136$\pm$0.005           & 0.243$\pm$0.019           & 62.451$\pm$0.122          & 77.868$\pm$0.350          & 83.216$\pm$0.416          \\
IMV-tensor & 0.061$\pm$0.013           & 0.103$\pm$0.009           & 0.187$\pm$0.009           & 60.579$\pm$0.233          & 77.621$\pm$0.483          & 81.979$\pm$0.596          \\ \hline
ETN-ODE    & \textbf{0.039$\pm$0.003}  & \textbf{0.094$\pm$0.004}  & \textbf{0.171$\pm$0.016}  & \textbf{59.909$\pm$0.599} & \textbf{76.543$\pm$0.361} & \textbf{80.694$\pm$0.657} \\ \hline
Dataset    & \multicolumn{3}{c|}{PM2.5}                                            & \multicolumn{3}{c}{Nasdaq100}                                         \\ \hline
TLSTM      & 68.346$\pm$0.145          & 71.845$\pm$0.242          & 76.169$\pm$0.571          & 74.804$\pm$0.220          & 76.037$\pm$1.615          & 75.260$\pm$0.459          \\
Latent-ODE & 41.062$\pm$6.494          & 50.461$\pm$4.521          & 61.723$\pm$1.481          & 33.750$\pm$4.637          & 36.256$\pm$2.728          & 33.033$\pm$7.921          \\
Retain     & 19.197$\pm$0.053          & 38.457$\pm$0.138          & 51.034$\pm$0.324          & 24.870$\pm$5.368          & 29.267$\pm$4.138          & 34.322$\pm$4.945          \\
UA         & 18.923$\pm$0.050          & 36.977$\pm$0.678          & 49.295$\pm$0.294          & \textbf{3.148$\pm$0.433}  & 5.555$\pm$1.131           & 8.352$\pm$0.586           \\
IMV-tensor & 17.694$\pm$0.031          & 35.553$\pm$0.370          & 48.058$\pm$0.175          & 12.821$\pm$3.031          & 18.831$\pm$4.084          & 22.001$\pm$6.886          \\ \hline
ETN-ODE    & \textbf{17.484$\pm$0.202} & \textbf{35.095$\pm$0.321} & \textbf{47.765$\pm$0.353} & 3.804$\pm$0.909           & \textbf{4.495$\pm$0.299}  & \textbf{6.661$\pm$1.254}  \\ \hline\hline
\end{tabular}
\caption{RMSE comparison on three different multi-step prediction tasks over the four datasets.}
\label{mtresult}
\end{table*}
We evaluate the various methods on five different standard multi-step prediction tasks, forecasting the next $1$,$2$, $5$, $8$, and $10$ future values of the target series. Due to the page limitation, the results of the $K=2, 8$, and MAE metrics are omitted here and are provided in the supplementary material.
It is observed that the proposed ETN-ODE achieves overall the best performance w.r.t. RMSE in Table~\ref{mtresult} demonstrating the effectiveness of the explainable continuous network.

In particular, ETN-ODE attains most significant improvements on SML2010 over the IMV-tensor baseline, generating an average $17.79\%$ decrease in RMSE.
On Nasdaq100, our model outperforms UA in the long-term forecasting tasks, while slightly inferior to UA on the short-term forecasting task ($K=1$). UA is originally designed for classification, which has advantages in short-term forecasting for multi-step prediction. ETN-ODE significantly outperforms another continuous-time version model, Latent-ODE. The hidden states are generated with a basic RNN network, which has limits in providing appropriate mappings to the ODE net when processing  multivariate time series in Latent-ODE. This result indicates the advantages of ETN-ODE in that  the TGRU layer produces informative hidden states representing different dynamics specific to each input feature. Moreover, the tandem attention layer helps to allocate different contributions to the predictions.

In a short summary, while ETN-ODE is designed mainly for continuous prediction, it could also lead to remarkable performance on standard multi-step time series prediction. Such advantages may be due to its better representation by parameterizing the derivative of the latent states.

\subsubsection{Ablation Study}
We design four variants to demonstrate the effectiveness of our model components: 
\begin{itemize}
\item \textbf{TLstm-ODE}: Replace the Tensorized GRU with Tensorized LSTM in ETN-ODE framework. 
\item \textbf{w/oODE}: Remove the ODE network component such that the model is no longer a continuous network. We use a fully connected layer to output the prediction.
\item \textbf{w/oATT}: Remove the tandem attention layer and use the mean pooling operated on the hidden states produced by the TGRU layer to generate the context vector $\bm{C_T}$.
\item \textbf{ODE-va}: Extract the variable attention in the tandem attention layer and obtain the prediction with a weighted sum of variable attention and latent states $\bm{z}$. 
\end{itemize}
\begin{figure}[h] 
	\centering
		\includegraphics[width=0.47\textwidth]{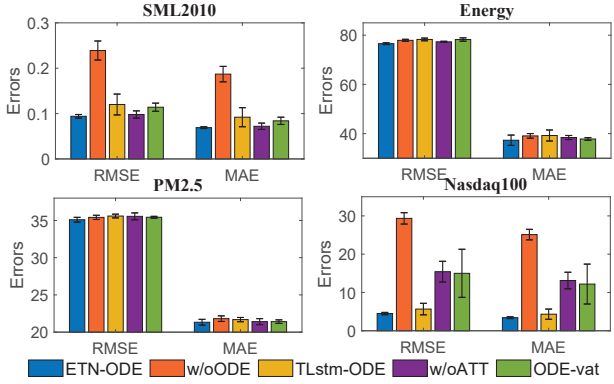}
	\caption{Results of ablation study.}
	\label{ablation}
\end{figure}

Taking the task of forecasting the next $5$ future values as an example,  we show in Fig.~\ref{ablation}  the ablation study. In general, with all the necessary components, ETN-ODE achieves the best performance consistently on the four datasets on both the metrics. In another word, removing any component from ETN-ODE could decline the model performance. Some important observations are highlighted as follows: 
\begin{itemize}
\item Most importantly, without ODE, the model could not output future values at arbitrary time points. It is efficient to model the gradient of latent states to improve performance as well.
\item Tensorized GRU appears more useful in attaining higher performance than Tensorized LSTM. We attribute the superiority to its fewer parameters, leading to faster convergence and better performance.
\item The attention component could substantially affect the prediction performance, which is crucial for ETN-ODE. The engaged tandem attention can also lead to the interpretability, which  will be visualized shortly.
\item Compared with ODE-va, the way of sampling the context vector $\bm{C_T}$ also influences the performance. Generating $\bm{C_T}$  with tandem attention could achieve better performance, which provides a better input to the ODE network. Adding attention after the ODE network hardly affects the model performance while modeling the gradient of latent states.
\end{itemize}

\subsubsection{Visualization of Tandem Attention}
\begin{figure}[b] 
	\centering  
	\subfigure[SML2010]{
		\label{v-att.sub1}
		\includegraphics[width=0.215\textwidth]{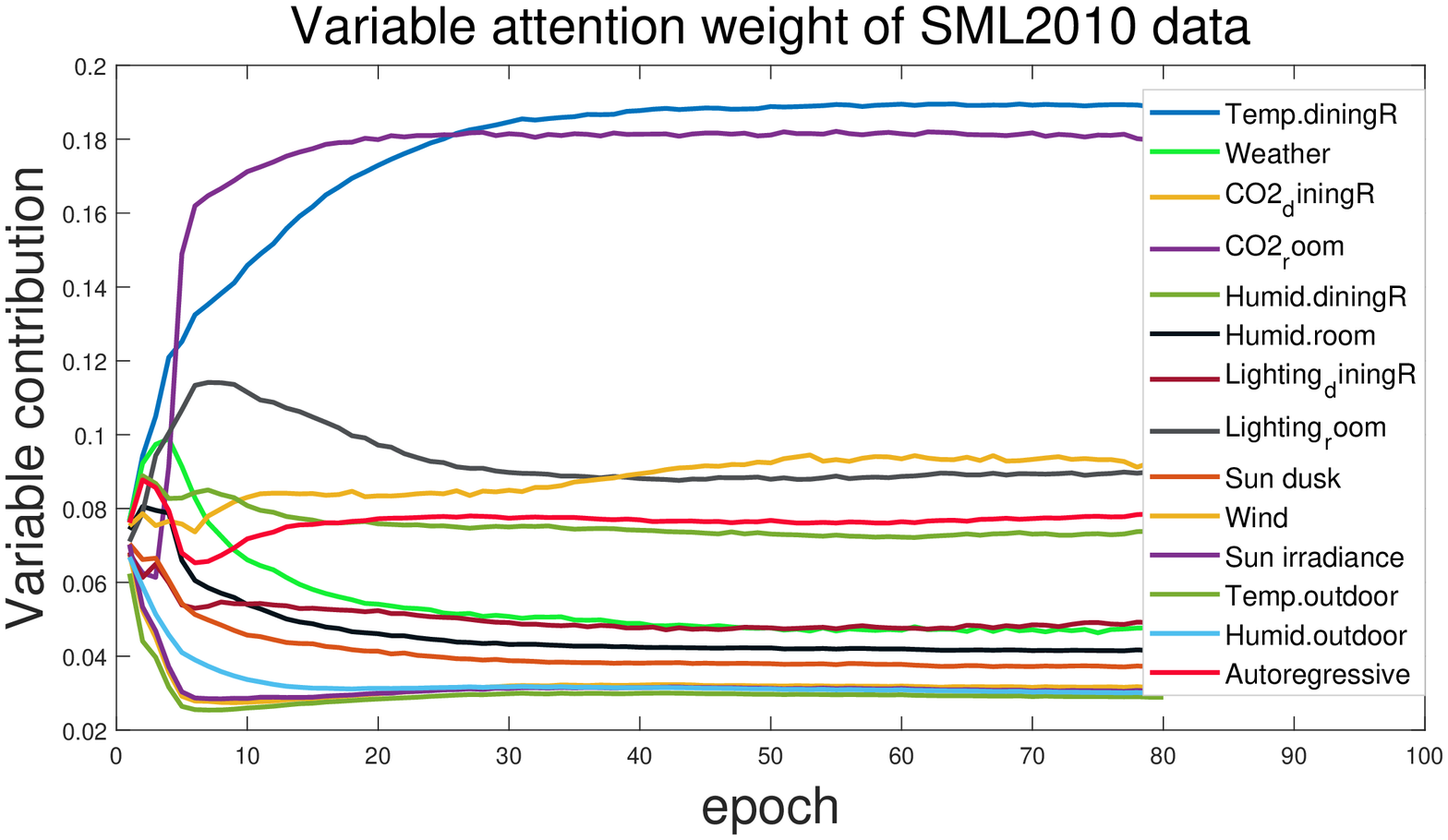}}
	\quad 
	\subfigure[PM2.5]{
		\label{v-att.sub2}
		\includegraphics[width=0.215\textwidth]{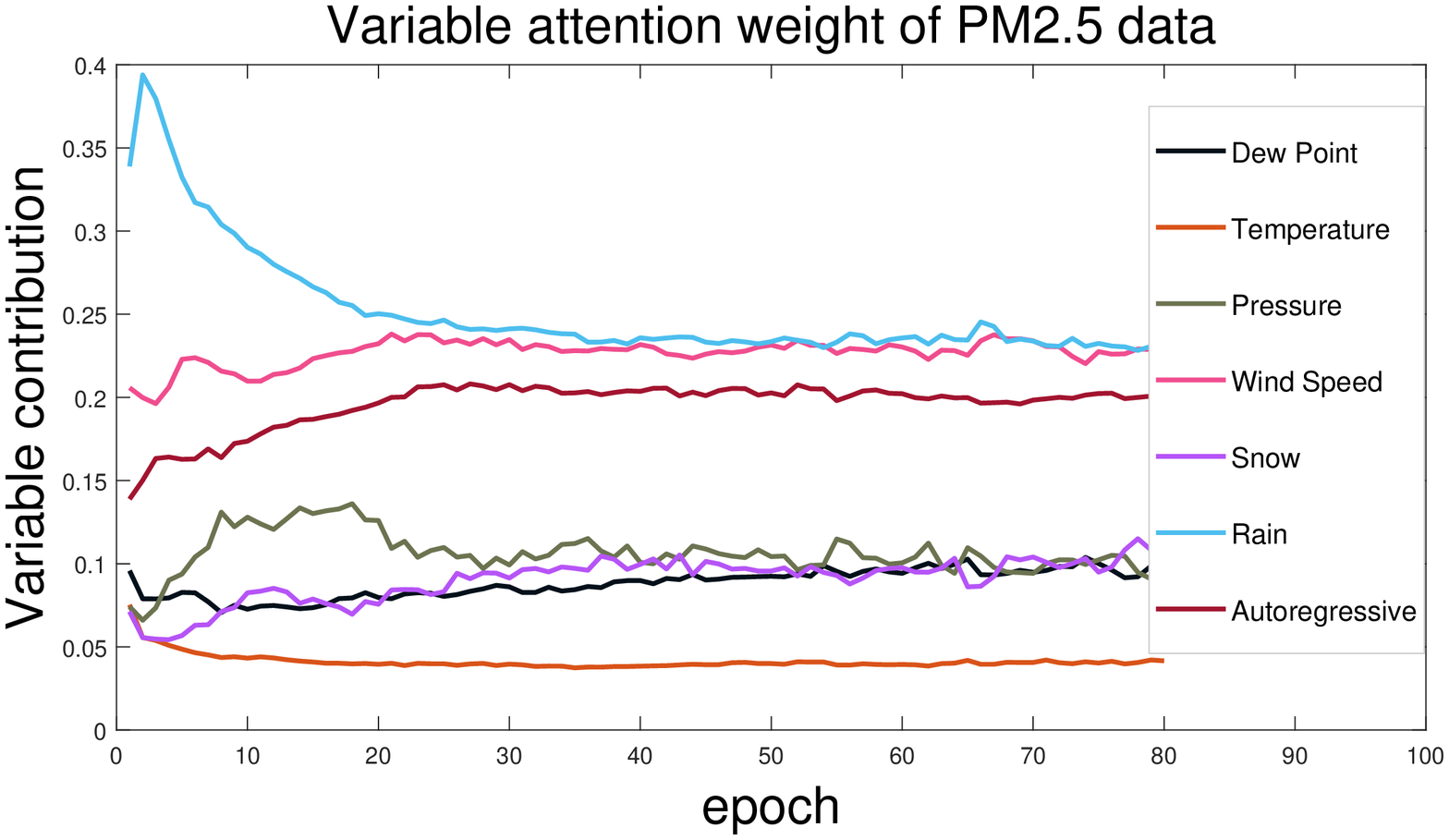}}
	\caption{Variable contribution on SML2010 and PM2.5.} 
	\label{v-att}
\end{figure}

With the tandem attention used in ETN-ODE, both the variable  and temporal contribution can be visualized, offering insights on the model interpretability. Such variable contribution is shown in Fig.~\ref{v-att} where variables with higher values contribute more to the predictions.
Specifically, in Fig.~\ref{v-att.sub1}, for SML2010 dataset, temperature in the dining room and Carbon dioxide content in the room have the most significant impact to the predictions. The temperature might rise when people cook in the kitchen. 
The attention mechanism is essential in our model, which could help to focus on more important driving series consistent with life experience. 

\begin{figure}[h] 
	\centering
		\includegraphics[width=0.47\textwidth]{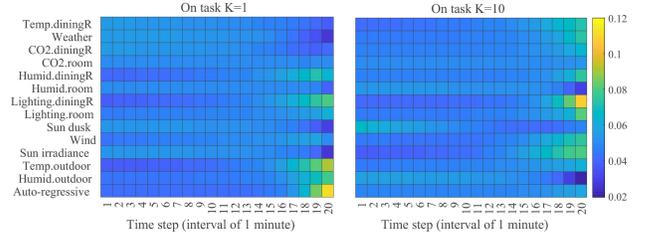}
	\caption{Temporal contribution on SML2010 at short/long-term forecasting ($K=1$ or $10$) at the end of training.}
	\label{t-att}
\end{figure}

In Fig.~\ref{t-att}, we demonstrate the temporal attention values at the end of training of SML2010 on short/long-term forecasting periods ($K=1$ or $10$). The lighter color represents that the corresponding data contributes more to the predictions. Specifically, short history of variables `Auto-regressive' and `Temp.outdoor' contributes more on the short-term forecasting period while lightings and sun irradiance influence more on the long-term forecasting period. We could also see that the temporal attention tends to allocate more weights to the long-historical data on the long-term forecasting period task.
Extra visualization and analysis of other datasets are presented in the supplementary material.


\subsubsection{Parameter Sensitivity}
\begin{figure}[h]
    \centering
    \subfigure{
    \label{sensitive.sub1}
    \includegraphics[width=0.143\textwidth]{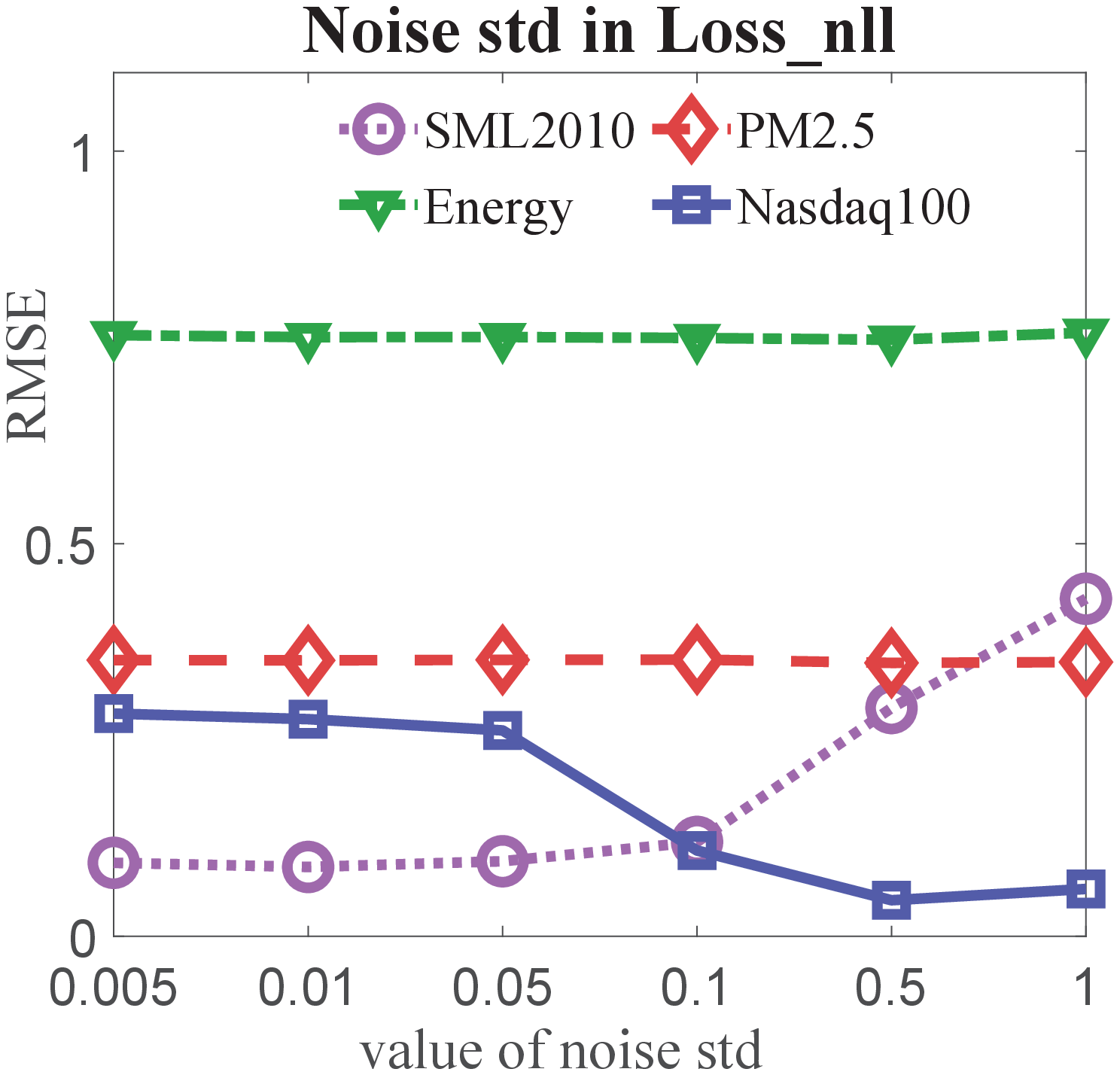}}
    \subfigure{
    \label{sensitive.sub2}
    \includegraphics[width=0.143\textwidth]{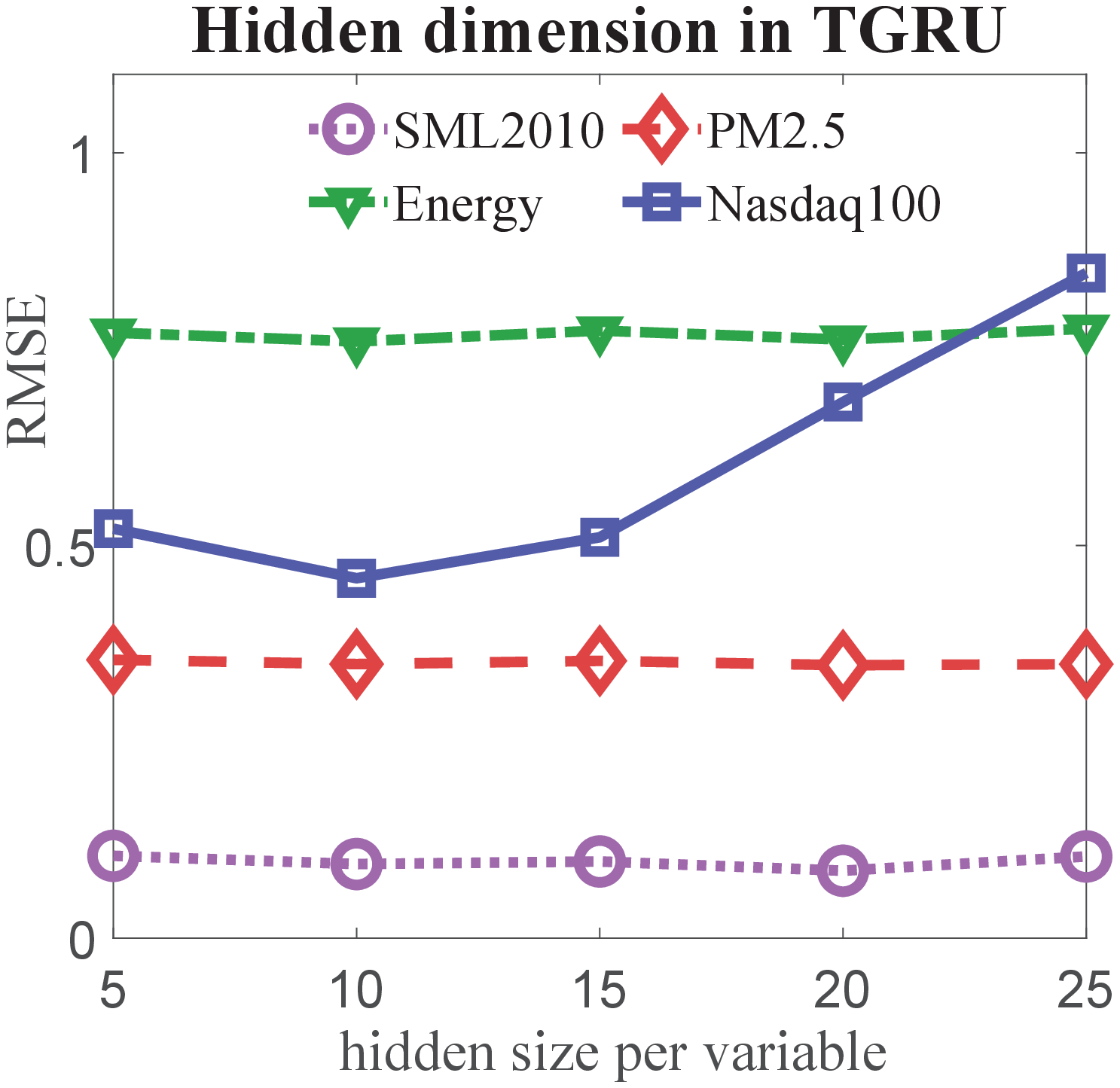}}
    \subfigure{
    \label{sensitive.sub3}
    \includegraphics[width=0.143\textwidth]{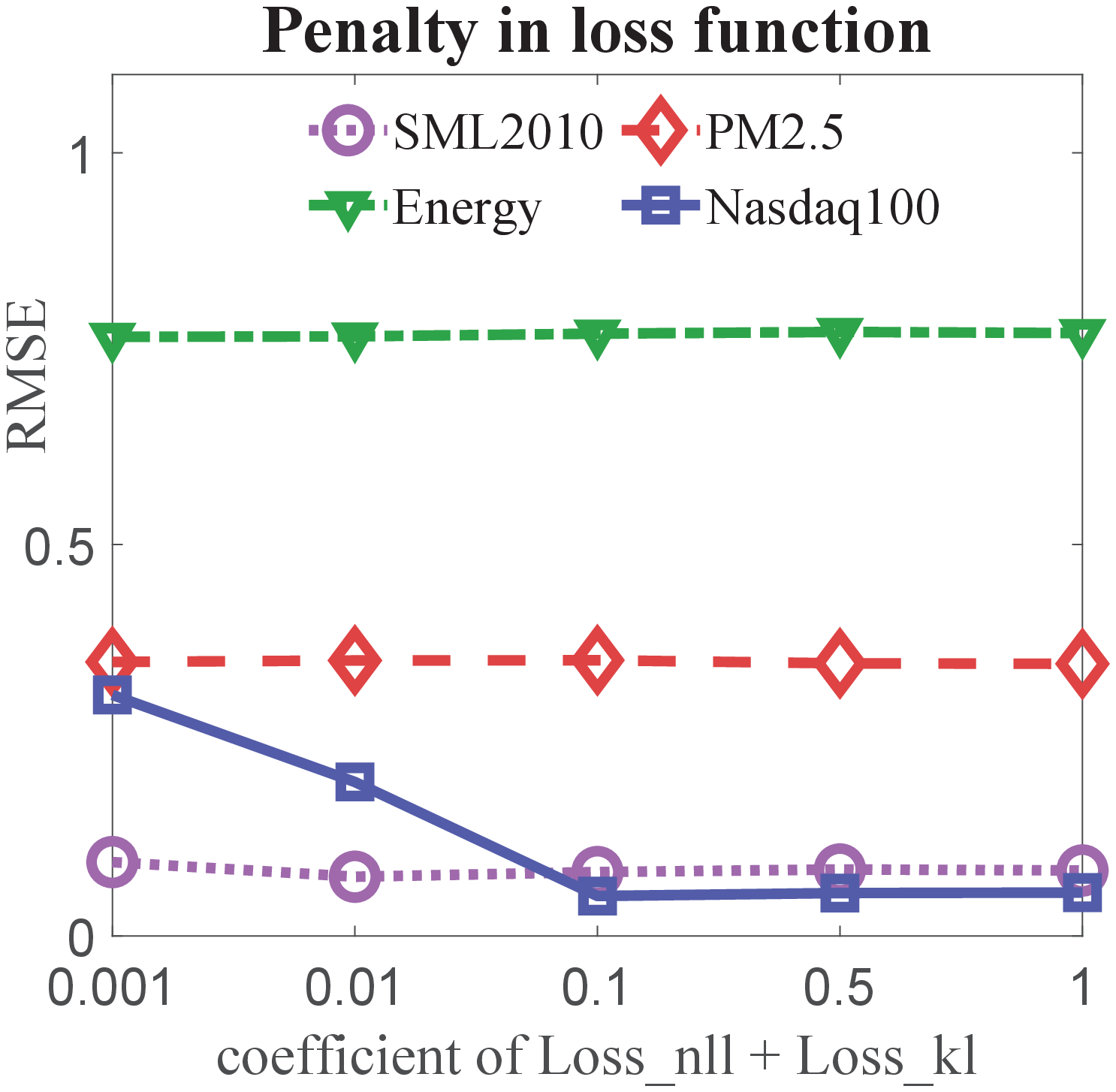}}
    \caption{Parameter sensitivity analysis ($K=5$).}
    \label{sensitive}
\end{figure}
Furthermore, we study the parameter sensitivity of ETN-ODE in Fig.~\ref{sensitive} w.r.t. the noise standard deviation (noise std.) in $Loss_{nll}$, hidden dimension per variable, and penalty in the loss function. We scale the RMSE error for better visualization on task $K=5$. For SML2010, the model performance would decline only when noise std. is greater than $0.05$. For Nasdaq100, it is relatively sensitive to the hyper-parameters mainly due to the large number of input features. For Energy and PM2.5, these parameters have limited influence on model performance. Overall, our model is not sensitive to hyper-parameters.

\section{Conclusion}
In this paper, we propose an explainable continuous framework (ETN-ODE) for arbitrary-step prediction of multivariate time series. The proposed ETN-ODE outputs arbitrary predicted values by utilizing an ODE network. We design a TGRU to process the multivariate time series representing different dynamics of individual input features with fewer parameters learned in networks. A tandem attention is designed to generate more adaptive inputs to the ODE network, providing interpretability by visualizing the temporal and variable contribution. Various experiments on arbitrary-step prediction and standard multi-step prediction on four real-world datasets demonstrate the effectiveness of our model.



\bibliography{AAAI}
\end{document}